\documentclass[pdflatex,sn-basic,Numbered,iicol]{sn-jnl}
\usepackage[utf8]{inputenc}
\usepackage[T1]{fontenc}
\usepackage{lmodern}
\usepackage[nopatch=footnote]{microtype}
\usepackage{amsmath,amssymb}
\usepackage{setspace}
\usepackage{graphicx,booktabs,makecell,tabularx}
\usepackage{xcolor}
\hypersetup{hidelinks,pdftitle={Synthetic reaction data from transformation rules for graph-to-sequence pretraining in biodegradation pathway~prediction},pdfauthor={Irsyaad Rijwan; Liam Brydon; Aidan Jared; J{\"o}rg Simon Wicker}}
\makeatletter\g@addto@macro\@maketitle{\vskip-36pt}\makeatother
\begin{document}
\title[enviGraph]{Synthetic reaction data from transformation rules for graph-to-sequence pretraining in biodegradation pathway~prediction}
\author*[1]{\fnm{Irsyaad} \sur{Rijwan}}\email{irij533@aucklanduni.ac.nz}
\author[1,2]{\fnm{Liam} \sur{Brydon}}
\author[1]{\fnm{Aidan} \sur{Jared}}
\author[1,2]{\fnm{J{\"o}rg Simon} \sur{Wicker}}
\affil[1]{\orgdiv{School of Computer Science}, \orgname{University of Auckland}, \orgaddress{\city{Auckland}, \country{New Zealand}}}
\affil[2]{\orgname{enviPath Limited}, \orgaddress{\city{Auckland}, \country{New Zealand}}}
\abstract{The scarcity of observed biodegradation reactions constrains the training of product and pathway predictors. We present enviGraph, a framework that generates and curates synthetic reaction data, uses these data to pretrain a graph-to-sequence predictor, and fine-tunes the predictor on observed environmental reactions for product and recursive pathway prediction. Applying inherited environmental transformation rules to structures from PubChem, NORMAN SusDat and EPA DSSTox, followed by curation, produces 776,851 canonical parent--product pairs for pretraining.

We compare this dataset with three United States Patent and Trademark Office (USPTO) reaction datasets: USPTO-50K, USPTO-MIT and USPTO-Stereo. We evaluate enviGraph and the sequence-based enviFormer using ten folds split by pathway on the EAWAG Biocatalysis/Biodegradation Database (BBD), regulatory pesticide biodegradation data (Soil) and activated-sludge biodegradation data (Sludge). For both predictors, synthetic pretraining gives the highest BBD mean precision--recall area under the curve (PR-AUC) among the evaluated pretraining datasets for single-generation and multi-generation prediction. enviGraph achieves BBD single-generation PR-AUC of $0.202\pm0.050$ with synthetic pretraining, compared with $0.169\pm0.065$ for the retrospectively selected strongest USPTO condition, USPTO-Stereo (mean $\pm$ sample standard deviation across folds). Its exact BBD top-1 and top-5 accuracies with synthetic pretraining are 0.355 and 0.594. The strongest USPTO conditions give higher PR-AUC means than synthetic pretraining in most Soil and Sludge comparisons. enviGraph has higher Soil multi-generation PR-AUC means than enviFormer under each of the four main pretraining datasets, and with synthetic pretraining it also leads enviFormer on BBD single-generation PR-AUC and exact top-1 and top-5 accuracy.

\noindent\textbf{Scientific Contribution.} We introduce enviGraph, a framework that generates and curates rule-derived synthetic reaction data for graph-to-sequence pretraining, followed by fine-tuning for biodegradation product and pathway prediction. The resulting curated corpus contains 776,851 canonical depth-1 parent--product pairs. Evaluation against USPTO pretraining with two predictors across BBD, Soil and Sludge, in both single-generation and multi-generation tasks, shows that the usefulness of the synthetic data varies across environmental datasets.}
\maketitle
\clearpage

\section{Introduction}
Organic contaminants can form transformation products with persistence, mobility and biological effects that differ from those of their parent compounds. The disappearance of a parent compound therefore provides an incomplete account of its environmental fate. Identifying transformation products experimentally is difficult when reference standards are unavailable or analytical methods do not capture short-lived intermediates. Computational prediction can help prioritise candidate structures and pathways for investigation, provided that predicted structures are distinguished from experimentally confirmed products~\cite{escher2011transformation,zahn2024transformation,schymanski2014confidence}.

Environmental reaction prediction faces a shortage of observed training examples. The enviPath resource provides compounds, reactions, pathways and transformation rules for microbial biodegradation, including the records inherited from the EAWAG Biocatalysis/Biodegradation Database (BBD), as well as Soil and Sludge collections~\cite{wicker2016envipath,wicker2024advancements}. These resources describe different environmental settings and experimental records. They are substantially smaller than the reaction datasets extracted from United States Patent and Trademark Office (USPTO) records~\cite{lowe2012uspto,lowe2017figshare} that are commonly used to train neural networks for reaction prediction.

Transformation rules encode reusable local changes that can be applied to many structures. Rule-based systems identify applicable transformations and construct candidate products; their coverage depends on the available rules and their specificity. Recursive application can produce a large number of candidate pathways. Earlier work combined transformation rules with learned prioritisation to control this expansion~\cite{fenner2008relative,wicker2010hybrid}. More recently, enviRule automated the extraction and generalisation of reaction patterns and evaluated them through downstream product prediction~\cite{zhang2023envirule}. A rule-generated product remains a hypothesis about a reaction's environmental occurrence. Using these products as training targets requires explicit choices about input structures, representation, acceptance, duplicate handling and evaluation-pair exclusion.

Neural networks can learn to generate reaction products from molecular inputs. Transformer models formulate product prediction as translation between molecular strings~\cite{schwaller2019molecular}. Pretrained initialisation improves generalisation over random initialisation in deep networks~\cite{erhan2010pretraining,yosinski2014transferable}. In reaction prediction, a USPTO-pretrained Molecular Transformer fine-tuned on about 20,000 carbohydrate reactions reached 70.3\% top-1 accuracy, compared with 30.4\% for a model trained only on those reactions~\cite{pesciullesi2020transfer}. Pretrained Chemformer models likewise outperformed randomly initialised counterparts on reaction and retrosynthesis prediction~\cite{irwin2022chemformer}.

In the environmental setting, enviFormer pretrains on USPTO reactions before adaptation to biodegradation data and evaluates both immediate products and recursively generated pathways~\cite{brydon2025enviformer}. Its authors concluded that the model's ability to predict products depends on transfer learning. This transfer strategy addresses the limited size of environmental datasets, but the USPTO examples describe intended chemical syntheses rather than transformation under environmental conditions. Whether rule-derived environmental examples provide a more useful pretraining dataset is an empirical question, and the answer may differ across target collections.

Recent approaches differ in their pretraining data, prediction inputs and output tasks. ReactionT5 combines compound pretraining with reaction pretraining on Open Reaction Database records, then evaluates product prediction, retrosynthesis and yield prediction, including limited-data settings~\cite{sagawa2025reactiont5}. Enzyformer combines molecular and reaction pretraining for enzymatic retrosynthesis~\cite{liu2026enzyformer}. Template-driven generation of training reactions has precedents in synthesis prediction. RSGPT, a transformer for retrosynthesis, was pretrained on more than 10 billion reactions generated by applying templates extracted from reported reactions to molecular fragments~\cite{deng2025rsgpt}. Grandjean et al.~\cite{grandjean2026data} applied retrosynthesis templates to USPTO molecules, validated 27.5 million fictive reactions and found that a retrosynthesis model trained on a template-equilibrated subset outperformed one trained on USPTO reactions only. These evaluations concern synthetic chemistry and enzymatic retrosynthesis. To our knowledge, no earlier study builds a pretraining corpus from environmental transformation rules, and our study tests such a corpus for environmental products and pathways. ECLIPSE predicts enzyme-class associations from compounds and evaluates their use as additional inputs to product prediction~\cite{brydonbrown2026eclipse}. In enviGraph, transformation rules supply training pairs, while prediction requires only the parent structure.

Molecular representation is a second design choice. A molecular graph directly exposes atom connectivity to an encoder, whereas a SMILES (Simplified Molecular Input Line Entry System) input specifies a traversal of that graph. Graph2SMILES combines local directed message passing and global graph-aware attention with sequence generation, providing a graph-to-sequence architecture for reaction prediction and retrosynthesis~\cite{tu2022graph2smiles}. UGTformer also combines graph pretraining with metabolic prediction, but its outputs are substrate and site-of-metabolism labels rather than product structures~\cite{zhan2026ugtformer}. We use an encoder derived from Graph2SMILES and an autoregressive product-SMILES decoder for environmental prediction. Comparing this predictor with enviFormer sets graph-to-sequence against sequence-to-sequence prediction on the same environmental data.

We introduce \emph{enviGraph} through three connected contributions:
\begin{enumerate}
\item We connect the generation and curation of rule-derived environmental reaction examples to graph-to-sequence pretraining, fine-tuning on observed reactions and biodegradation product and pathway prediction.
\item We construct and document a curated corpus of 776,851 canonical depth-1 parent--product pairs, including its seed sources, transformation-rule provenance and curation procedure.
\item We compare synthetic and USPTO pretraining for enviGraph and enviFormer across BBD, Soil and Sludge in single-generation and multi-generation tasks. We also compare the predictors under the same pretraining datasets. Synthetic pretraining gives the highest BBD mean precision--recall area under the curve (PR-AUC) for both predictors; the strongest USPTO conditions lead most Soil and Sludge comparisons.
\end{enumerate}
Rule execution and neural optimisation remain separate stages within the framework. Multi-generation evaluation follows the procedure used for enviFormer and established work on missing intermediates and multi-step recovery~\cite{brydon2025enviformer,tam2021holistic}.

\section{The enviGraph framework}
\label{sec:framework}

\begin{figure*}[!tp]
\centering
\includegraphics[width=\textwidth]{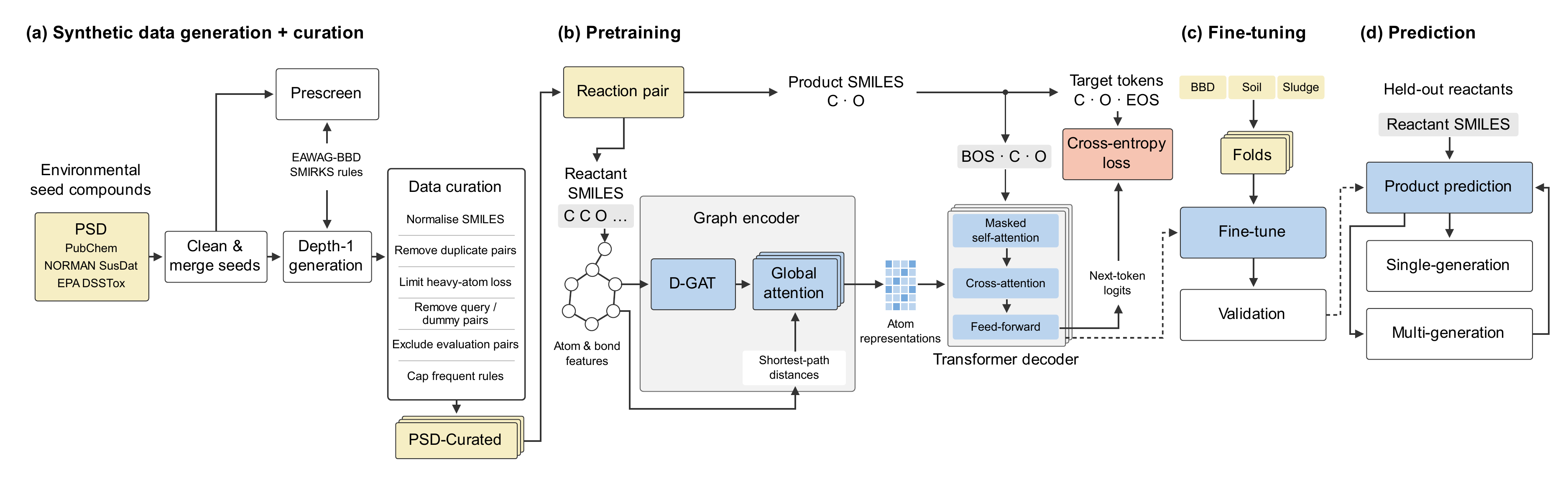}
\caption{The integrated enviGraph framework.}
\label{fig:workflow}
\par\small\raggedright
(a) Seed structures and inherited transformation rules produce depth-1 synthetic reactions. Prescreening measures valid-product rule coverage as a diagnostic. Curation standardises and filters pairs, excludes evaluation-pair matches and caps frequent rules. Additional file 1 describes the smaller PS seed mixture and its corpora. (b) A parent--product pair supplies a molecular graph and a tokenised product. A directed graph attention network (D-GAT), followed by global attention using shortest-path distances in bonds, produces atom representations for decoder cross-attention. Shifted product tokens supply decoder inputs and next-token targets, with beginning-of-sequence (BOS) and end-of-sequence (EOS) markers. Cross-entropy trains the graph encoder and decoder jointly. The molecular sketch is schematic. (c) Pretrained weights initialise models that are fine-tuned on the training portion of each pathway-aware fold; internal validation selects the checkpoint. (d) Held-out parents are decoded into candidate products. Recursive expansion applies the selected predictor to newly encountered products and is evaluated using multi-generation PR-AUC. Solid arrows show computational flow; dashed arrows show parameter transfer. Seed inventory details, curation stages, search settings and metric definitions are given in the text. The full-size vector figure accompanies the manuscript for zoomed viewing.
\end{figure*}

Figure~\ref{fig:workflow} connects the four stages of enviGraph: synthetic reaction generation and curation (a), graph-to-sequence pretraining (b), fine-tuning on observed environmental reactions (c), and prediction (d). Stage (a) exports parent--product pairs for pretraining. We write each training reaction as a reaction SMILES string, \texttt{reactant>{}>product}. In USPTO reactions, the reactant side can list several reactants and reagents. In the generated pairs and in the BBD, Soil and Sludge pathways, it is a single parent compound. During both training stages, the reactant supplies the molecular graph and the product supplies shifted decoder inputs and next-token targets. Fine-tuning adapts the pretrained weights to each environmental training partition.

At inference, the selected predictor receives only the parent structure and returns scored product candidates. Newly predicted products can become inputs for recursive pathway expansion. Rule identity and environmental scenario are retained in construction records where available, but are not conditioning inputs to the predictor.

\subsection{Transformation rules and seed structures}

We obtained transformation rules from the EAWAG-BBD package in enviPath \cite{wicker2016envipath}. The retained rule set contained 498 top-level rule objects and 578 extracted transformation entries; exact-string deduplication produced 289 executable SMIRKS entries. These counts are not interchangeable because composite and simple rules can expose the same transformation. We compiled the cached strings using RDKit's \texttt{ReactionFromSmarts} \cite{rdkit}. Compilation establishes executability under RDKit's reaction-SMARTS (SMILES arbitrary target specification) semantics, which differ from the original Daylight SMIRKS specification \cite{daylightsmirks}.

Seed structures came from PubChem, NORMAN SusDat and the United States Environmental Protection Agency (EPA) Distributed Structure-Searchable Toxicity (DSSTox) database \cite{kim2023pubchem,taha2022norman,williams2017comptox}. Cleaning selected the largest carbon-containing fragment, applied an organic-element filter and canonicalised SMILES, without a separate charge-neutralisation or tautomer-canonicalisation step. Merging used stored or recomputed Standard InChIKeys (hashed International Chemical Identifiers), with canonical SMILES as a fallback \cite{heller2015inchi}. Source identifiers and aggregate provenance were retained.

The principal seed mixture, PSD, combines 14,047 cleaned source records from PubChem keyword searches for pharmaceutical, industrial-chemical and pollutant terms, 96,070 from NORMAN SusDat, 1,056,537 from DSSTox and 2,029 from a PubChem keyword search for pesticides. Its 1,168,683 cleaned source records yield 1,090,572 unique seeds. Additional file 1 describes PS, a smaller mixture without DSSTox and the pesticide search.

\subsection{Generating and curating synthetic reactions}

Generation applies each compiled transformation rule to each cleaned seed molecule unless a cached negative reactant filter matches, producing candidate parent--product records. Before generation, reaction-mode prescreening measures valid-product rule coverage as a diagnostic; it does not restrict which rules generation applies.

Each generated product is sanitised with RDKit and reduced to its largest carbon-containing fragment. Generation rejects sanitisation failures, excluded elements, negative product-filter matches, identity products and products with fewer than two heavy atoms. Duplicate canonical products are removed within each parent--rule application. When a rule returns several product molecules, each surviving molecule is recorded separately. A parent--product pair therefore need not describe a complete stoichiometric reaction.

The evaluated synthetic corpora use depth~1: rules act on seed molecules without recursively expanding their products. Generation records preserve the parent, product, rule, SMIRKS, generation depth and available origin metadata. Exported model input files contain canonical \texttt{parent>{}>product} pairs shared by the graph and sequence predictors. The generated products are weak training labels, without experimental confirmation of their environmental occurrence, rate or yield.

Curation begins by clearing isotope labels, removing redundant explicit hydrogens, and writing non-isomeric canonical SMILES. This policy also removes stereochemical distinctions. Dropping isotope-labelled rows instead would have removed 55 of the 134 represented SMIRKS entries while discarding only about 0.14\% of rows. Failed and identity pairs are discarded, and stable pair deduplication retains one rule provenance per example. We then remove pairs losing more than 45 heavy atoms and pairs containing query-bond or dummy-atom tokens. The heavy-atom loss threshold follows the observed 99th percentile of loss magnitude in the PSD construction audit. Gains remain unbounded because a corresponding gain threshold would exclude conjugation families.

We exclude synthetic pairs matching evaluation reactions from BBD, Soil or Sludge. The overlap key selects the largest-heavy-atom reactant and product after fragment selection and canonicalises their isotope-normalised, non-isomeric pair. This exclusion removes direct matches under that reduced representation; it does not separate substrates or scaffolds and does not remove the transformations that the BBD reactions share with the EAWAG-BBD rules.

Rule-frequency capping follows evaluation-pair exclusion, so removing matched pairs cannot leave a capped rule below its cap. Among the 1,935,577 normalised unique pairs, the five, ten and twenty most frequent rules accounted for 56.14\%, 74.25\% and 91.95\% of pairs. To limit this concentration, PSD-Curated retains at most 36,526 pairs per SMIRKS, the rounded-up 90th percentile of per-SMIRKS counts among these pairs. Stable sorting by rule, canonical structures and identifiers makes retention deterministic. Capping reduces frequent-rule exposure without oversampling rare rules. The thresholds were selected from corpus diagnostics, and the experiments compare the complete corpora that curation produces.

\subsection{Graph-to-sequence product prediction}

The enviGraph predictor combines a Graph2SMILES-derived molecular encoder \cite{tu2022graph2smiles} with an autoregressive product-SMILES decoder using Transformer-XL-style relative self-attention \cite{dai2019transformerxl}. We use the name enviGraph for both the framework and its trained predictor. We adapted these inherited mechanisms to the environmental corpus, training and pathway interfaces. The encoder first aggregates local atom and bond information, then uses global attention to relate atoms across the molecule. The decoder reads these atom representations while generating the product SMILES one token at a time.

Each molecular bond becomes two directed edges. Atom features describe element, degree, formal charge, hybridisation, valence, hydrogen count, chirality and aromaticity; bond features describe type, stereochemistry, conjugation and ring membership. The directed graph attention network (D-GAT) uses four shared attention and gated-update iterations to aggregate incoming bond messages while excluding the immediate reverse edge. A subsequent atom-level aggregation produces 256-dimensional states.

Six global attention layers then combine these atom states using bucketed shortest-path distances measured in bonds. Each layer has eight attention heads and feed-forward width 2,048. Distances zero through seven have separate buckets; eight through fourteen share one, and fifteen or more share another. Disconnected pairs have their own bucket.

Examples that fail graph parsing or SMILES tokenisation are omitted.

The six-layer decoder has dimension 256, eight heads and feed-forward width 2,048. Masked self-attention reads earlier product tokens, and cross-attention reads the contextual atom representations. Absolute sinusoidal input positions accompany relative self-attention with signed offset $\operatorname{clip}(j-i,-4,4)$. Clipping groups distant offsets; it does not restrict attention to four tokens. The decoder has no recurrence between sequence segments.

For reactant $x$, graph $G(x)$ and product token sequence $y$, both pretraining and fine-tuning minimise
\begin{equation}
\begin{aligned}
\mathcal{L}(\theta)={}&-\sum_{(x,y)\in\mathcal{D}}\sum_{t=1}^{|y|}
\\[-2pt]&\qquad\log p_{\theta}(y_t\mid y_{<t},G(x)).
\end{aligned}
\end{equation}
Teacher forcing supplies the preceding reference product tokens, and minibatch loss averages over non-padding target tokens, including the end-of-sequence token. At inference, the decoder instead uses its preceding predicted tokens. The reference configuration uses zero label smoothing.

\section{Experimental design}
\label{sec:experiments}

\subsection{Datasets and baseline}

We evaluate two types of comparison: pretraining conditions within each predictor, and the graph and sequence predictors under the same pretraining dataset. The main pretraining datasets are PSD-Curated, USPTO-50K~\cite{schneider2016reactionroles}, USPTO-MIT~\cite{jin2017wl} and USPTO-Stereo~\cite{schwaller2018translation}. We also compare PSD-Curated with two corpora from the PS seed mixture (Section~\ref{sec:additional-corpora}). The profiled PSD-Curated file contains 776,851 unique valid canonical reactions.

We compare enviGraph with enviFormer, the environmental sequence-to-sequence predictor introduced by Brydon et al.\ \cite{brydon2025enviformer}. We pretrain both predictors using the same pretraining datasets. Within-predictor comparisons assess complete pretraining conditions; comparisons using the same pretraining dataset set graph-to-sequence against sequence-to-sequence prediction, each predictor in its reference configuration. The reference enviFormer configuration uses dimension 256, eight heads, four encoder and four decoder layers, and feed-forward width 2,048. Approximate reference parameter counts are 18.2 million for enviGraph and 11.7 million for enviFormer.

BBD, Soil and Sludge are evaluated as separate datasets. BBD contains curated biodegradation pathways; Soil originates from regulatory pesticide studies; Sludge records activated-sludge transformations \cite{wicker2016envipath,latino2017soil,trostel2023predictive}. Table~\ref{tab:paper-targets} describes retained local package objects before fold construction. BBD is the one dataset from the same EAWAG-BBD package as the transformation rules used in synthetic generation.

The object counts of the local packages differ from the deduplicated reaction populations obtained after pathway preprocessing and from the subsets scored in multi-generation evaluation. We therefore use the package counts to describe the data resources and use the saved pathway-fold records to identify evaluation partitions.

\begin{table*}[!tp]
\centering
\caption{Object counts in the local environmental packages before fold construction.}
\label{tab:paper-targets}
\begin{tabular}{lrrrr}
\hline
Dataset & Compounds & Reactions & Rules & Pathways \\
\hline
BBD & 1,399 & 1,480 & 290 & 219 \\
Soil & 2,609 & 2,447 & 80 & 317 \\
Sludge & 1,070 & 521 & 0 & 184 \\
\hline
\end{tabular}
\end{table*}

\subsection{Splits and training}

Pretraining normally groups reactions by reactant into 80/10/10 partitions. For downstream evaluation, we construct ten pathway-aware folds by shuffling whole pathways with split seed~1. Preprocessing bypasses pseudo-nodes, canonicalises node structures, extracts edge-derived pairs and omits carbon-dioxide-only products. Reaction strings are deduplicated separately within training and test partitions. For each fold, a training reaction is excluded if its reactant equals a test reactant or its entire product string equals a test product. This preserves pathway allocation while allowing related structures and some cross-side overlaps. Within the remaining training set, a reactant-grouped 90/10 split with seed~1 supplies internal validation; the training-framework seed is~2.

Each downstream fold starts from pretrained weights and updates all trainable parameters. Both predictors use Adam \cite{kingma2015adam} and Transformer-style warm-up \cite{vaswani2017attention}. We select checkpoints using whole-sequence validation accuracy under teacher-forced argmax decoding. For fresh fine-tuning, enviGraph creates a new optimiser but offsets its learning-rate schedule by the global step reached during pretraining. In the implementation evaluated here, enviFormer restarts its learning-rate schedule during fresh fine-tuning, whereas the published enviFormer study continued the pretraining schedule~\cite{brydon2025enviformer}.

Batch size and learning-rate scale differed between pretraining runs but were the same for all fine-tuning runs, and dropout differed between predictors.

We conducted the experiments using NVIDIA RTX PRO 6000 and L4 GPUs.

\subsection{Single-generation prediction and evaluation}

We assess single-generation predictions in two ways. Precision--recall evaluation counts recovery when a candidate contains all components of a reference product. Exact top-$k$ accuracy requires equality between the predicted and reference component sets. Both use candidates returned by beam decoding, scored by raw sequence log-probability,
\begin{equation}
s_\theta(y\mid x)=\sum_t\log p_\theta(y_t\mid y_{<t},x).
\end{equation}
Partial sequences are retained during beam search using length normalisation \cite{wu2016gnmt}, with $\alpha=0$ for enviGraph and $-0.5$ for enviFormer. Completed candidates are ranked and thresholded using their raw scores. Reference single-generation beam widths are five and eight, respectively. These scores depend on string length and are not calibrated occurrence probabilities.

For precision--recall evaluation, we run inference for each reference reaction row, then pool candidates by canonical parent. The evaluator deduplicates product-component sets and retains the first stored score for duplicate candidates. It counts recovered reference rows and unique parent--candidate pairs separately. Let $\mathcal{X}$ be the distinct test parents, $R_j$ the reference component set in row $j$ for parent $x$, and $m_x$ its number of reference rows. For unique predictions $\widehat{\mathcal{P}}_\tau(x)$ satisfying $s_\theta>\tau$, we calculate
\begin{equation}
\begin{aligned}
H(\tau)={}&\sum_{x\in\mathcal{X}}\sum_{j=1}^{m_x}
\\[-2pt]&\mathbf{1}\left[\exists\widehat{R}\in\widehat{\mathcal{P}}_\tau(x):R_j\subseteq\widehat{R}\right],
\end{aligned}
\end{equation}
\begin{equation}
\begin{aligned}
C(\tau)&=\sum_x|\widehat{\mathcal{P}}_\tau(x)|,
\\ P(\tau)&=\frac{H(\tau)}{C(\tau)},
\qquad R(\tau)=\frac{H(\tau)}{\sum_xm_x},
\end{aligned}
\end{equation}
with $P=0$ for an empty candidate set. Recall counts recovered reference rows, whereas the precision denominator counts unique parent--candidate pairs. One candidate can recover several reference rows, so this precision is a reference-recovery ratio and is not intrinsically bounded by one.

Exact top-$k$ accuracy instead counts each distinct parent once: success requires equality between a reference component set and one of its first $k$ non-empty score-ranked saved predictions. We report top-1 and top-5, including parents whose complete returned lists are empty or shorter than the requested rank in the denominator. Product-component order does not affect either metric, but precision--recall containment and top-$k$ equality define different recovery criteria. The published enviFormer evaluation used containment for top-$k$ matching~\cite{brydon2025enviformer}; our exact, parent-level measure therefore should not be compared directly with its reported top-$k$ values.

\subsection{Multi-generation prediction and evaluation}

We use the multi-generation biodegradation pathway evaluation employed by Brydon et al.~\cite{brydon2025enviformer} for enviFormer, which builds on the depth weighting and intermediate-product treatment introduced by Tam et al.~\cite{tam2021holistic}. Starting from an initial compound, the predictor generates products and uses newly encountered products as inputs for subsequent generations. This recursive expansion constructs a predicted pathway graph. Evaluation prunes the graph at a series of score thresholds and compares its retained compounds with the reference pathway using the depth-weighted recovery procedure below.

We evaluate reference pathways containing exactly one initial compound, defined as a depth-0 node. A predicted node first reached from parent $u$ receives cumulative score $S(v)=S(u)+s_\theta(y_v\mid u)$, with root score zero; components of the same string share its sequence score. Rediscovery adds an edge without replacing the first stored score.

Reference search settings request five beams per molecule and retain at most 50 product strings per layer by local sequence score. The nominal depth argument is seven, but a stored reference pathway depth can override it. If $D$ is the argument after that override, generation runs through $D-1$. Expansion also stops at an empty frontier or after a completed layer exceeds 1,000 nodes.

At each threshold, pruning removes cycles by deleting edges, then removes nodes below threshold and non-root nodes disconnected from the root. Following Tam et al.\ \cite{tam2021holistic}, compound recovery is weighted by $2^{-d}$, where $d$ denotes depth. Intermediates are internal nodes on a shortest predicted path between recovered reference neighbours; identified intermediates absent from the reference receive zero weight. Other predicted nodes use depths adjusted by subtracting identified intermediates along a selected shortest root path. True-positive and false-negative weights use reference depths; false-positive weights use adjusted predicted depths. Roots are excluded. This measures weighted compound recovery without requiring every edge to match.

We calculate precision and recall for each pathway at each threshold, then average each metric across the evaluated pathways within a fold. Varying the threshold produces one precision--recall curve per fold, whose area gives its multi-generation PR-AUC. Pathway averaging therefore occurs within each fold before the fold-level areas are summarised.

For both single-generation and multi-generation evaluation, the threshold grid spans $-20.0$ to $-2.1$ by 0.1 and $-2.00$ to $-0.01$ by 0.01. PR-AUC uses trapezoidal integration after stable ascending-recall ordering, without added endpoints or a precision envelope. For both tasks, reported values are the mean and sample standard deviation of ten fold-specific PR-AUC values. Shared training data make folds dependent; these dispersions describe partition variation, not independent optimisation repetitions.

Fine-tuned performance is the main analysis. Comparisons between pretraining datasets use the highest eligible mean among USPTO-50K, USPTO-MIT and USPTO-Stereo, selected retrospectively for each predictor, evaluation dataset and prediction task. Summaries of the highest-scoring predictor also select pretraining datasets retrospectively; comparisons using the same pretraining dataset fix it for both predictors. Because every model is pretrained before fine-tuning, the comparisons concern the choice of pretraining dataset rather than whether to pretrain.

\section{Results}
\label{sec:results}

\subsection{Construction and curation of the synthetic corpus}

The generation component of enviGraph applied environmental transformation rules to xenobiotic structure inventories to produce pretraining pairs for the graph and sequence predictors (Fig.~\ref{fig:workflow}). The PSD seed mixture contained 1,090,572 unique seed structures after merging 1,168,683 cleaned source records. Generation produced 2,003,522 depth-1 rows. The complete curation sequence retained 776,851 depth-1 parent--product pairs, or 38.8\% of the generated rows, spanning 120 represented SMIRKS entries (Table~\ref{tab:corpus}). Representation processing and pair deduplication removed 67,945 rows, the heavy-atom loss filter 12,521, the query-bond and dummy-atom filter 96,438 and evaluation-pair exclusion 426. The rule cap removed a further 1,049,341 rows, and the share assigned to the five most frequent rules decreased from 56.14\% to 23.51\%.

The corpus profile identified 337,512 distinct reactants, 550,977 distinct products and 81,283 non-empty product scaffolds in PSD-Curated. Its export contained no unparsable rows and no query-bond or dummy-atom tokens, and no additional matches were found under the recorded exact-pair overlap check. The PSD-Uncurated file contained 1,934,327 unique valid reactions after profiling; it was characterised but was not used in a predictive-performance comparison. Curation reduced molecular and scaffold counts. The top-ten scaffold share among ring-bearing unique products increased from 22.12\% to 29.78\% for PSD, showing that reduced rule concentration did not imply increased scaffold diversity. The two profiles describe different-sized outputs rather than size-controlled diversity gains.

\begin{table*}[!tp]
\centering
\caption{Construction and curation of the PSD synthetic corpus.}
\label{tab:corpus}
\small
\setlength{\tabcolsep}{5pt}
\renewcommand{\arraystretch}{1.08}
\begin{tabular}{lr}
\toprule
PSD construction population & Count \\
\midrule
Cleaned source records & 1,168,683 \\
Unique seed structures & 1,090,572 \\
Represented SMIRKS entries, before / after curation & 134 / 120 \\
Top-five rule share, before / after curation & 56.14\% / 23.51\% \\
\bottomrule
\end{tabular}
\par\vspace{0.7em}
\begin{tabular}{lrrr}
\toprule
Curation stage & Rows & Removed & Retained (\%) \\
\midrule
Generated depth-1 rows & 2,003,522 & --- & 100.0 \\
Representation processing and pair deduplication & 1,935,577 & 67,945 & 96.6 \\
Heavy-atom loss filter (loss above 45) & 1,923,056 & 12,521 & 96.0 \\
Query-bond and dummy-atom filter & 1,826,618 & 96,438 & 91.2 \\
Evaluation-pair exclusion & 1,826,192 & 426 & 91.1 \\
Rule cap (36,526 pairs per SMIRKS) & 776,851 & 1,049,341 & 38.8 \\
\bottomrule
\end{tabular}
\par\vspace{0.5em}
\begin{minipage}{\textwidth}\footnotesize
Each stage removes the listed rows from the output of the previous stage; retention is relative to the generated depth-1 rows. From the second stage onward, each row is a normalised unique parent--product pair. Rule shares use separate rankings before and after the complete curation sequence. File profiles of the curated and uncurated corpora are given in Additional file 1: Table S1.
\end{minipage}
\end{table*}

\subsection{Comparison of synthetic and USPTO pretraining}

Table~\ref{tab:prauc} compares pretraining conditions within each predictor for single-generation and multi-generation PR-AUC on BBD, Soil and Sludge. Results are means and sample standard deviations across ten pathway-aware folds, with one execution per fold. Among the four main pretraining datasets, PSD-Curated gave both predictors their highest BBD means in both tasks.

\begin{figure*}[!tp]
\centering
\includegraphics{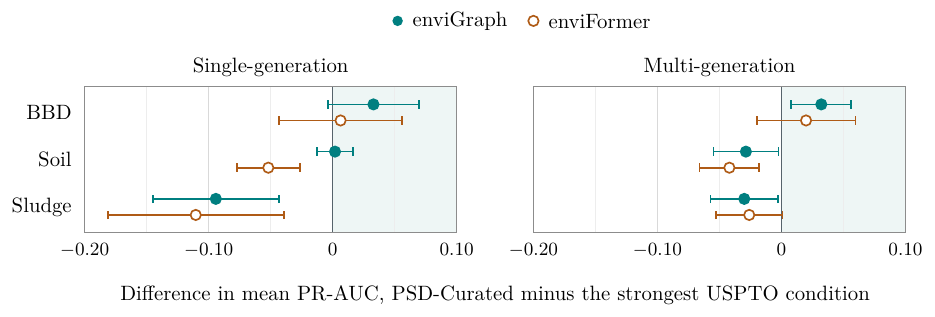}
\caption{PR-AUC differences between synthetic and USPTO pretraining.}
\label{fig:source-deltas}
\par\small\raggedright
Positive differences favour PSD-Curated. Each USPTO value is the highest mean among USPTO-50K, USPTO-MIT and USPTO-Stereo for that predictor, evaluation dataset and prediction task, selected retrospectively. Points are mean differences over the ten pathway-aware folds; error bars show $\pm1$ sample standard deviation of the ten paired fold differences. The folds share training data, so the error bars are not confidence intervals.
\end{figure*}

On BBD, enviGraph pretrained on PSD-Curated achieved single-generation PR-AUC of $0.202\pm0.050$, compared with $0.169\pm0.065$ for USPTO-Stereo, its strongest USPTO condition for this task. Its multi-generation PR-AUC increased from $0.078\pm0.028$ with USPTO-Stereo to $0.110\pm0.029$ with PSD-Curated. The corresponding synthetic-minus-USPTO differences were 0.033 and 0.032 (Fig.~\ref{fig:source-deltas}). enviFormer also had higher BBD means with synthetic pretraining: $0.175\pm0.054$ for single-generation prediction and $0.112\pm0.037$ for multi-generation prediction, exceeding the USPTO-MIT and USPTO-Stereo conditions by 0.006 and 0.020, respectively. The BBD differences were positive for both predictors, with larger differences for enviGraph.

PSD-Curated did not consistently outperform USPTO pretraining on Soil and Sludge. For enviGraph on Soil, PSD-Curated was slightly above USPTO-Stereo in single-generation prediction, 0.154 versus 0.152, but below it in multi-generation prediction, 0.119 versus 0.148. enviFormer achieved higher Soil PR-AUC with USPTO-Stereo in both tasks. On Sludge, the highest USPTO means exceeded PSD-Curated for both predictors and tasks: USPTO-MIT gave the highest single-generation means and USPTO-Stereo the highest multi-generation means. The complete values and fold dispersions are reported in Table~\ref{tab:prauc}.

The USPTO condition in each pretraining comparison was selected retrospectively from USPTO-50K, USPTO-MIT and USPTO-Stereo. Paired tests between pretraining datasets gave adjusted $q$-values of 0.0496 and 0.0139 for the two positive enviGraph BBD differences; the corresponding enviFormer values were 0.6853 and 0.3690. Benjamini--Hochberg adjustment was applied to a broader set of tests between pretraining datasets, of which these comparisons are a subset. Because pretraining-dataset selection was post hoc and the cross-validation training sets overlap, these tests are exploratory.

Multi-generation PR-AUC was lower than single-generation PR-AUC in all 24 combinations of predictor, main pretraining dataset and evaluation dataset. The closest means were obtained by enviGraph/USPTO-Stereo on Soil, at 0.148 for multi-generation and 0.152 for single-generation prediction.

\begin{table*}[!tp]
\centering
\caption{Fine-tuned single-generation and multi-generation PR-AUC.}
\label{tab:prauc}
\footnotesize
\setlength{\tabcolsep}{2.8pt}
\renewcommand{\arraystretch}{1.05}
\resizebox{\textwidth}{!}{%
\begin{tabular}{llcccccc}
\toprule
& & \multicolumn{2}{c}{BBD} & \multicolumn{2}{c}{Soil} & \multicolumn{2}{c}{Sludge} \\
\cmidrule(lr){3-4}\cmidrule(lr){5-6}\cmidrule(lr){7-8}
Predictor & Pretraining & \makecell{Single\\PR-AUC\\$\uparrow$} & \makecell{Multi\\PR-AUC\\$\uparrow$} & \makecell{Single\\PR-AUC\\$\uparrow$} & \makecell{Multi\\PR-AUC\\$\uparrow$} & \makecell{Single\\PR-AUC\\$\uparrow$} & \makecell{Multi\\PR-AUC\\$\uparrow$} \\
\midrule
enviGraph & PSD-Curated & \textbf{\makecell{0.202\\$\pm$ 0.050}} & \makecell{0.110\\$\pm$ 0.029} & \makecell{0.154\\$\pm$ 0.037} & \makecell{0.119\\$\pm$ 0.035} & \makecell{0.017\\$\pm$ 0.017} & \makecell{0.010\\$\pm$ 0.004} \\
enviFormer & PSD-Curated & \makecell{0.175\\$\pm$ 0.054} & \textbf{\makecell{0.112\\$\pm$ 0.037}} & \makecell{0.130\\$\pm$ 0.028} & \makecell{0.088\\$\pm$ 0.027} & \makecell{0.044\\$\pm$ 0.045} & \makecell{0.018\\$\pm$ 0.021} \\
\addlinespace[4pt]
enviGraph & USPTO-50K & \makecell{0.107\\$\pm$ 0.027} & \makecell{0.052\\$\pm$ 0.017} & \makecell{0.103\\$\pm$ 0.024} & \makecell{0.098\\$\pm$ 0.046} & \makecell{0.043\\$\pm$ 0.033} & \makecell{0.019\\$\pm$ 0.014} \\
enviFormer & USPTO-50K & \makecell{0.111\\$\pm$ 0.034} & \makecell{0.051\\$\pm$ 0.024} & \makecell{0.106\\$\pm$ 0.019} & \makecell{0.071\\$\pm$ 0.022} & \makecell{0.060\\$\pm$ 0.060} & \makecell{0.014\\$\pm$ 0.010} \\
\addlinespace[4pt]
enviGraph & USPTO-MIT & \makecell{0.148\\$\pm$ 0.048} & \makecell{0.068\\$\pm$ 0.025} & \makecell{0.149\\$\pm$ 0.038} & \makecell{0.126\\$\pm$ 0.057} & \makecell{0.112\\$\pm$ 0.049} & \makecell{0.037\\$\pm$ 0.015} \\
enviFormer & USPTO-MIT & \makecell{0.169\\$\pm$ 0.043} & \makecell{0.089\\$\pm$ 0.035} & \makecell{0.175\\$\pm$ 0.032} & \makecell{0.116\\$\pm$ 0.029} & \textbf{\makecell{0.154\\$\pm$ 0.061}} & \makecell{0.038\\$\pm$ 0.017} \\
\addlinespace[4pt]
enviGraph & USPTO-Stereo & \makecell{0.169\\$\pm$ 0.065} & \makecell{0.078\\$\pm$ 0.028} & \makecell{0.152\\$\pm$ 0.032} & \textbf{\makecell{0.148\\$\pm$ 0.053}} & \makecell{0.061\\$\pm$ 0.037} & \makecell{0.040\\$\pm$ 0.027} \\
enviFormer & USPTO-Stereo & \makecell{0.163\\$\pm$ 0.052} & \makecell{0.092\\$\pm$ 0.047} & \textbf{\makecell{0.182\\$\pm$ 0.039}} & \makecell{0.130\\$\pm$ 0.038} & \makecell{0.137\\$\pm$ 0.070} & \textbf{\makecell{0.044\\$\pm$ 0.022}} \\
\bottomrule
\end{tabular}}
\par\vspace{0.5em}
\begin{minipage}{\textwidth}\footnotesize
Means and sample standard deviations over ten pathway-aware folds. Single and Multi denote single-generation PR-AUC and multi-generation PR-AUC. Bold identifies the largest unrounded mean across all eight rows in each column; emphasis is descriptive. Results for the PS corpora are in Additional file 1: Table S2.
\end{minipage}
\end{table*}

\subsection{Comparison of the graph and sequence predictors}

We compared the predictors within each pretraining dataset, evaluation dataset and prediction task. Across all four main pretraining datasets, enviGraph had the higher Soil multi-generation mean, whereas enviFormer had the higher Sludge single-generation mean. BBD single-generation rankings changed with the pretraining dataset: enviGraph led under PSD-Curated and USPTO-Stereo, while enviFormer led under USPTO-50K and USPTO-MIT. With PSD-Curated, enviGraph exceeded enviFormer by 0.027 on BBD single-generation PR-AUC, 0.023 on Soil single-generation PR-AUC and 0.031 on Soil multi-generation PR-AUC, using unrounded means. enviFormer was higher on both Sludge tasks and by 0.002 on BBD multi-generation PR-AUC. Across the four main pretraining datasets, enviGraph led six of the twelve multi-generation comparisons, against three of the twelve single-generation comparisons.

\subsection{Exact top-\texorpdfstring{$k$}{k} accuracy}

Exact top-$k$ accuracy provides a stricter product-matching complement to PR-AUC (Table~\ref{tab:topk}). On BBD, enviGraph with PSD-Curated achieved the highest top-1 accuracy, $0.355\pm0.052$, and top-5 accuracy, $0.594\pm0.077$. Relative to enviFormer with the same pretraining dataset, the differences were 0.060 and 0.064. The BBD advantage of enviGraph with PSD-Curated was therefore also observed when recovery required an exact match to a reference component set.

For Soil, enviFormer/USPTO-Stereo led top-1 and top-5 accuracy at $0.332\pm0.051$ and $0.610\pm0.051$. The highest Sludge accuracies came from different pretraining datasets: enviFormer/USPTO-MIT reached top-1 accuracy of $0.345\pm0.099$, while enviFormer/USPTO-Stereo reached top-5 accuracy of $0.577\pm0.128$. All reported top-$k$ conditions were evaluated over ten folds. These results measure exact-reference recovery within ranked candidate lists; they do not assess the chemical validity of additional, unrecorded products.

\begin{table*}[!tp]
\centering
\caption{Fine-tuned exact top-1 and top-5 accuracy.}
\label{tab:topk}
\footnotesize
\setlength{\tabcolsep}{2.8pt}
\renewcommand{\arraystretch}{1.05}
\resizebox{\textwidth}{!}{%
\begin{tabular}{llcccccc}
\toprule
& & \multicolumn{2}{c}{BBD} & \multicolumn{2}{c}{Soil} & \multicolumn{2}{c}{Sludge} \\
\cmidrule(lr){3-4}\cmidrule(lr){5-6}\cmidrule(lr){7-8}
Predictor & Pretraining & Top-1 $\uparrow$ & Top-5 $\uparrow$ & Top-1 $\uparrow$ & Top-5 $\uparrow$ & Top-1 $\uparrow$ & Top-5 $\uparrow$ \\
\midrule
enviGraph & PSD-Curated & \textbf{\makecell{0.355\\$\pm$ 0.052}} & \textbf{\makecell{0.594\\$\pm$ 0.077}} & \makecell{0.316\\$\pm$ 0.046} & \makecell{0.563\\$\pm$ 0.070} & \makecell{0.063\\$\pm$ 0.035} & \makecell{0.263\\$\pm$ 0.083} \\
enviFormer & PSD-Curated & \makecell{0.295\\$\pm$ 0.057} & \makecell{0.530\\$\pm$ 0.056} & \makecell{0.268\\$\pm$ 0.050} & \makecell{0.525\\$\pm$ 0.063} & \makecell{0.108\\$\pm$ 0.087} & \makecell{0.344\\$\pm$ 0.109} \\
\addlinespace[4pt]
enviGraph & USPTO-50K & \makecell{0.246\\$\pm$ 0.032} & \makecell{0.461\\$\pm$ 0.053} & \makecell{0.248\\$\pm$ 0.052} & \makecell{0.509\\$\pm$ 0.045} & \makecell{0.125\\$\pm$ 0.055} & \makecell{0.379\\$\pm$ 0.139} \\
enviFormer & USPTO-50K & \makecell{0.229\\$\pm$ 0.047} & \makecell{0.396\\$\pm$ 0.069} & \makecell{0.243\\$\pm$ 0.042} & \makecell{0.475\\$\pm$ 0.058} & \makecell{0.161\\$\pm$ 0.099} & \makecell{0.351\\$\pm$ 0.150} \\
\addlinespace[4pt]
enviGraph & USPTO-MIT & \makecell{0.282\\$\pm$ 0.063} & \makecell{0.482\\$\pm$ 0.059} & \makecell{0.308\\$\pm$ 0.052} & \makecell{0.580\\$\pm$ 0.070} & \makecell{0.252\\$\pm$ 0.052} & \makecell{0.492\\$\pm$ 0.114} \\
enviFormer & USPTO-MIT & \makecell{0.307\\$\pm$ 0.058} & \makecell{0.530\\$\pm$ 0.055} & \makecell{0.323\\$\pm$ 0.049} & \makecell{0.600\\$\pm$ 0.041} & \textbf{\makecell{0.345\\$\pm$ 0.099}} & \makecell{0.530\\$\pm$ 0.109} \\
\addlinespace[4pt]
enviGraph & USPTO-Stereo & \makecell{0.298\\$\pm$ 0.072} & \makecell{0.535\\$\pm$ 0.069} & \makecell{0.300\\$\pm$ 0.044} & \makecell{0.602\\$\pm$ 0.061} & \makecell{0.119\\$\pm$ 0.071} & \makecell{0.503\\$\pm$ 0.170} \\
enviFormer & USPTO-Stereo & \makecell{0.286\\$\pm$ 0.050} & \makecell{0.520\\$\pm$ 0.069} & \textbf{\makecell{0.332\\$\pm$ 0.051}} & \textbf{\makecell{0.610\\$\pm$ 0.051}} & \makecell{0.288\\$\pm$ 0.103} & \textbf{\makecell{0.577\\$\pm$ 0.128}} \\
\bottomrule
\end{tabular}}
\par\vspace{0.5em}
\begin{minipage}{\textwidth}\footnotesize
Means and sample standard deviations over ten pathway-aware folds. Accuracy is the fraction of parents for which one of the first one or five non-empty score-ranked predictions exactly matches a reference component set. Bold identifies the largest unrounded mean across all eight rows in each column; emphasis is descriptive. Results for the PS corpora are in Additional file 1: Table S3.
\end{minipage}
\end{table*}

\subsection{Comparison with a smaller seed mixture}
\label{sec:additional-corpora}

We also pretrained both predictors on PS-Curated and PS-Uncurated, two corpora generated from the smaller PS seed mixture of 103,079 unique seeds (Additional file 1: Tables S2 and S3). Across single-generation PR-AUC, multi-generation PR-AUC and exact top-1 and top-5 accuracy, PSD-Curated gave the highest mean of the three synthetic corpora in 20 of the 24 combinations of predictor, evaluation dataset and metric. It led all twelve enviFormer combinations and the eight enviGraph combinations on BBD and Soil. The four exceptions were enviGraph on Sludge, where both PS corpora gave higher means for every metric. Seed mixture, corpus size, rule cap and some fine-tuning settings changed together between PS and PSD, so the comparison supports PSD-Curated, built from the larger seed mixture, as the principal synthetic corpus.

\section{Discussion}
\label{sec:discussion}

\subsection{Rule-derived training data and corpus curation}

Earlier biodegradation systems applied rules directly to substrates and ranked the resulting products \cite{hou2003functional,ellis2008umpps,fenner2008relative}. In enviGraph, rule execution supplies synthetic reaction examples for pretraining; fine-tuning then adapts the predictor to observed environmental reactions. The contribution is the construction and evaluation of these training examples within a framework for biodegradation product and pathway prediction. The transformation rules remain those inherited from the EAWAG-BBD package in enviPath \cite{wicker2016envipath}.

Curation filtered and rebalanced the generated pairs before pretraining. Evaluation-pair exclusion removed direct matches to BBD, Soil and Sludge reactions under the reduced overlap key. The structural filters removed extreme fragmentation and query-bond or dummy-atom tokens, and rule capping limited exposure to a few frequent transformations (Table~\ref{tab:corpus}). Reduced rule concentration did not imply greater scaffold diversity, since the top-ten scaffold share increased during curation. In the PS mixture, the curated corpus gave lower PR-AUC means than its uncurated counterpart in eleven of the twelve combinations of predictor, evaluation dataset and prediction task, under training conditions that were not fully matched (Additional file 1: Table S2). A natural next experiment separates the effect of curation from that of corpus size.

The generated pairs expand the substrate contexts available for pretraining, while remaining limited to transformations in the rule library. They do not confirm environmental occurrence, specify reaction conditions or retain every reaction participant. The predictor receives only the parent structure. Its subsequent recovery of recorded products and pathways provides the test of whether these generated examples are useful for the evaluation datasets.

\subsection{Synthetic pretraining across BBD, Soil and Sludge}

PSD-Curated gave the highest BBD PR-AUC means for both predictors in both prediction tasks (Table~\ref{tab:prauc}). The shared direction across the graph and sequence predictors supports the usefulness of PSD-Curated for BBD, although the differences from USPTO were smaller for enviFormer. The shared origin of the generation rules and the BBD reactions offers a plausible explanation: both come from the EAWAG-BBD package, which supplied the transformations from which the cached rules were obtained. Brydon et al.~\cite{brydon2025enviformer} likewise found expert rules to be a strong BBD baseline. Evaluation-pair exclusion removes direct matches under the overlap key, but related scaffolds, motifs and transformation types can remain. The BBD result therefore supports transfer to reactions from the database that supplied the rules; it does not show the same advantage on Soil or Sludge, which come from other enviPath packages.

The BBD pretraining advantage did not extend consistently to Soil or Sludge (Fig.~\ref{fig:source-deltas}), so the pretraining condition with the highest mean depended on the evaluation dataset and prediction task. Chemical coverage, dataset size and training conditions differ between the pretraining datasets, and varying them one at a time would show which of them accounts for the Soil and Sludge results. Lower performance than USPTO pretraining is also not evidence that synthetic pretraining harms performance relative to random initialisation. The strongest USPTO conditions were selected retrospectively from the same cross-validation results; these comparisons remain exploratory.

\subsection{Predictor comparison and multi-generation prediction}

The Soil multi-generation results distinguish predictor performance from pretraining-dataset performance. enviGraph had higher means than enviFormer under all four main pretraining datasets, but its highest Soil multi-generation mean came from USPTO-Stereo: 0.148, compared with 0.119 for PSD-Curated. On BBD with PSD-Curated, enviGraph led enviFormer on single-generation PR-AUC, 0.202 against 0.175, and on exact top-1 and top-5 accuracy.

Multi-generation prediction depends on successive product predictions. Missing an intermediate can prevent recovery of its descendants, while retaining an incorrect intermediate can create additional branches. Single-generation and multi-generation PR-AUC use different matching and aggregation rules, so the lower multi-generation values (Table~\ref{tab:prauc}) are not a direct measure of error accumulation. Recall at each depth and the number of predicted branches per layer would show where multi-generation recovery is lost and help explain enviGraph's Soil multi-generation ranking. Reference pathways can also omit products and intermediates, so multi-generation PR-AUC measures recovery of recorded compounds under this evaluation procedure. Unmatched predictions are candidate products rather than confirmed errors, and every predicted product remains a hypothesis about environmental formation.

\section{Conclusions}
\label{sec:conclusions}

enviGraph integrates the construction of synthetic reaction data with graph-to-sequence pretraining and fine-tuning for biodegradation product and pathway prediction. Its principal curated corpus contains 776,851 depth-1 parent--product pairs generated from inherited environmental transformation rules and xenobiotic structures. Curation excluded pairs matching evaluation reactions and reduced the share of the five most frequent rules from 56.14\% to 23.51\%. This corpus gave higher means than both corpora from a smaller seed mixture in 20 of the 24 combinations of predictor, evaluation dataset and metric.

Among the evaluated conditions, PSD-Curated gave both predictors their highest BBD PR-AUC means in single-generation and multi-generation prediction. enviGraph pretrained on PSD-Curated also achieved the highest exact BBD top-1 and top-5 accuracy. The strongest USPTO conditions gave higher PR-AUC means than PSD-Curated in most Soil and Sludge comparisons. enviGraph had higher Soil multi-generation means than enviFormer under all four main pretraining datasets, with its best mean obtained using USPTO-Stereo. The strongest evidence for the synthetic corpus is therefore on BBD, where it also gave enviGraph its lead over enviFormer in single-generation PR-AUC and exact top-$k$ accuracy. The graph-to-sequence predictor's standing improved on the recursive task: it led six of the twelve multi-generation comparisons, against three of the twelve single-generation comparisons.

\section*{Abbreviations}
\begin{list}{}{\setlength{\labelwidth}{4.5em}\setlength{\labelsep}{1em}\setlength{\leftmargin}{5.5em}\setlength{\topsep}{0pt}\setlength{\itemsep}{0pt}\setlength{\parsep}{0pt}}\raggedright
\item[BBD\hfill] EAWAG Biocatalysis/Biodegradation Database
\item[BOS\hfill] Beginning of sequence
\item[D-GAT\hfill] Directed graph attention network
\item[DSSTox\hfill] Distributed Structure-Searchable Toxicity
\item[EOS\hfill] End of sequence
\item[EPA\hfill] United States Environmental Protection Agency
\item[GPU\hfill] Graphics processing unit
\item[InChIKey\hfill] Hashed International Chemical Identifier
\item[PR-AUC\hfill] Precision--recall area under the curve
\item[PS\hfill] PubChem and NORMAN SusDat seed mixture
\item[PSD\hfill] PubChem, NORMAN SusDat and DSSTox seed mixture
\item[SMARTS\hfill] SMILES arbitrary target specification
\item[SMILES\hfill] Simplified Molecular Input Line Entry System
\item[SMIRKS\hfill] Daylight reaction transform language
\item[USPTO\hfill] United States Patent and Trademark Office
\end{list}

\section*{Declarations}
\subsection*{Availability of data and materials}
The study uses the public resources described and cited in Sections~\ref{sec:framework} and~\ref{sec:experiments}. The seed structures used to generate the synthetic corpora, the trained checkpoints and the evaluation outputs will be made publicly available. The Python source code of enviGraph will be released under the GNU Affero General Public License v3.0 (AGPL-3.0).

\subsection*{Competing interests}
JSW is one of the founders of enviPath Limited, a scientific software development company that develops and maintains enviPath. LB is a Machine Learning Engineer at enviPath Limited. The authors declare no other competing interests.

\subsection*{Funding}
Not applicable.

\subsection*{Authors' contributions}
IR designed and implemented enviGraph, conducted the experiments and analysis, and drafted the manuscript. LB provided code and guidance for running the enviFormer experiments and evaluation, and reviewed and proofread the manuscript. AJ reviewed and proofread the manuscript. JSW supervised the research, provided expertise, guidance and support throughout, and reviewed and proofread the manuscript. All authors read and approved the final manuscript.

\subsection*{Acknowledgements}
Not applicable.

{\singlespacing

}

\section*{Additional files}
\noindent Additional file 1\\
File format: .pdf\\
Title: PS seed-mixture corpora and fine-tuned results\\
Description: Construction and curation of the corpora from the smaller PS seed mixture, with file profiles of the PS and PSD corpora (Section S1, Table S1); PR-AUC and exact top-1 and top-5 accuracy for PS-Curated, PS-Uncurated and PSD-Curated after fine-tuning (Section S2, Tables S2 and S3).

\end{document}